\documentclass[10pt,twocolumn,letterpaper]{article}

\usepackage{iccv}
\usepackage{times}
\usepackage{epsfig}
\usepackage{graphicx}
\usepackage{amsmath}
\usepackage{amssymb}
\usepackage{color}
\usepackage[normalem]{ulem}
\usepackage{pdfpages}

\usepackage[pagebackref=true,breaklinks=true,letterpaper=true,colorlinks,bookmarks=false]{hyperref}
\DeclareMathOperator*{\argmin}{arg\,min}

\iccvfinalcopy 

\def\iccvPaperID{6728} 

\ificcvfinal\fi

\begin{document}

\title{Indoor Scene Generation from a Collection of Semantic-Segmented Depth Images}

\author{{Ming-Jia Yang}\thanks{This work is done when Ming-Jia Yang was an intern at MSRA}\,\,$^{1,2}$\quad{Yu-Xiao Guo}$^{2}$\quad{Bin Zhou}$^{1}$\quad{Xin Tong}$^{2}$\\
$^1${Beihang University}\quad\quad$^2${Microsoft Research Asia}\\
{\tt\small\{yangmingjia,zhoubin\}@buaa.edu.cn\quad\{yuxgu,xtong\}@microsoft.com}
}

\maketitle
\ificcvfinal\thispagestyle{empty}\fi

\begin{abstract}      
We present a method for creating 3D indoor scenes with a generative model learned from a collection of semantic-segmented depth images captured from different unknown scenes. Given a room with a specified size, our method automatically generates 3D objects in a room from a randomly sampled latent code. Different from existing methods that represent an indoor scene with the type, location, and other properties of objects in the room and learn the scene layout from a collection of \textbf{complete 3D} indoor scenes, our method models each indoor scene as a 3D semantic scene volume and learns a volumetric generative adversarial network (GAN) from a collection of \textbf{2.5D partial} observations of 3D scenes. To this end, we apply a differentiable projection layer to project the generated 3D semantic scene volumes into semantic-segmented depth images and design a new multiple-view discriminator for learning the complete 3D scene volume from 2.5D semantic-segmented depth images. Compared to existing methods, our method not only efficiently reduces the workload of modeling and acquiring 3D scenes for training, but also produces better object shapes and their detailed layouts in the scene. We evaluate our method with different indoor scene datasets and demonstrate the advantages of our method. We also extend our method for generating 3D indoor scenes from semantic-segmented depth images inferred from RGB images of real scenes. \footnote{Code URL: \url{ https://github.com/mingjiayang/SGSDI}}
   
\end{abstract}

\section{Introduction}
Real-world indoor scenes exhibit rich variations with different numbers, types, and layouts of the objects placed in a room due to different interior designs and living activities. Generating realistic 3D indoor scenes is an important task for many applications, such as VR/AR, 3D game design, and robotic navigation. 

Manually modeling indoor scenes with variant and realistic object layouts in a room is a labor-intensive task and requires professional skills. Automatic scene generation techniques try to model the properties and distributions of the objects in real scenes and generate new 3D scenes in two steps. For a room with a specified size and shape, these methods first determine the layout (i.e. orientation and position) and properties (e.g. type and shape) of the objects in the room. After that, they retrieve a CAD model of each object from a 3D object database based on the object's properties and then place the resulting CAD models in the room according to their layout. 

A set of methods have been developed for modeling the properties and distributions of objects in indoor scenes. Early methods use manually defined rules \cite{merrell2011interactive} or simple statistic models \cite{yu2011make,fisher2012example,chang2015text,fu2017adaptive,ma2018language,qi2018human} computed from scene instances for generating a specific type of scenes, which are difficult to generalize to other types of scenes. Recent deep-learning-based methods \cite{li2019grains,zhang2020deep,wang2018deep,ritchie2019fast,wang2019planit} learn a deep neural network of the object properties and layouts from a large collection of 3D scene instances that are difficult to be modeled by skilled artists or captured from real scenes. By simply modeling the object geometry with their sizes, these methods fail to model concrete 3D object shapes and the detailed object layouts determined by their shapes, such as a chair with their seat under a desk or a TV inside a cabinet. 

In this paper, we present a generative adversarial network (GAN) for 3D indoor scene generation. Different from previous methods that represent the scene with object properties and layouts, our method models a 3D indoor scene with a semantic scene volume, where each voxel is either labeled as empty or the type of object that it belongs to. Based on this representation, we design a volumetric GAN model that takes the room size as input and synthesizes the semantic scene volumes of the room that consist of different objects and their layouts from randomly sampled latent vectors. After that, our method generates the final 3D indoor scene by replacing each volumetric object instance in the volume with a CAD model retrieved from a 3D object database based on their type and volumetric shape. 

Different from previous methods that train the networks with a collection of complete 3D indoor scenes, we learn the volumetric GAN model from a collection of semantic-segmented depth images, each of which captures a 2.5D partial view of an unknown 3D scene. To this end, we apply a differentiable projection layer between the generator and discriminator, which projects the generated semantic scene volume into semantic-segmented depth images from a set of views. We then feed both projected fake semantic-segmented depth images and real semantic-segmented depth images into the discriminator for GAN training. 

A naive design of the discriminator is to use the single-view discriminator for learning 3D object representation from 2D images \cite{nguyen2019hologan,ProGAN2017,li2019synthesizing}. Unfortunately, the GAN model trained with this single-view discriminator is prone to generating indoor scenes with unnatural object layouts. We thus propose a multi-view discriminator that takes a combination of multiple views rendered from generated scenes for GAN training. 
Since the training images are captured from different unknown scenes and we have no scene ID of each image, we use a random combination of training images to approximate the ground truth layouts of underlying scenes. For this purpose, we empirically figure out the optimal number and type of views of the random training image combinations that can well approximate the underlying scene layouts and facilitate the GAN training.

To the best of our knowledge, our method is the first approach that learns to generate 3D indoor scenes from a collection of semantic-segmented depth images, which greatly reduces the workload for training data acquisition and modeling. Thanks to semantic scene volume representation, our method can better model the object shapes and their detailed layouts than existing methods. We evaluate our method both synthetic Structured3D \cite{zheng2019structured3d} and real MatterPort3D \cite{chang2017matterport3d} datasets and demonstrate the advantages of our method. With the help of existing RGB2Depth methods, we show that our method can successfully learn 3D scene generation from segmented-depth images inferred from RGB images of real scenes.  


\section{Related work}
\paragraph{3D Indoor Scene Generation.} Early methods synthesize a specific type of 3D scene with manually defined rules \cite{merrell2011interactive,chang2015text}, simple statistic models learned from 3D scene instances \cite{yu2011make,fu2017adaptive,fisher2012example,ma2018language}, or And-Or Graph (AoG)\cite{qi2018human} of all valid object distributions in a scene. 


Learning-based methods model the object properties and layouts in 3D scenes with a deep neural network learned from an annotated 3D scene dataset. All these methods abstract the objects and their properties (e.g. position, orientation, type, and size) in a scene as nodes with attributes and represent the 3D scenes as top-view 2D images \cite{wang2018deep,ritchie2019fast}, node graphs \cite{wang2019planit,luo2020end,zhang2020deep} or trees \cite{li2019grains} in scene layout generation. All these methods require a collection of 3D annotated scenes for training. Moreover, most of these methods except \cite{zhang2020deep} represent the object geometry as its bounding box and thus fail to model the object shape and detailed object layouts. Different from these methods, we learn a GAN model from a collection of semantic-segmented depth images captured from the scenes. our method models the scene as a semantic scene volume and can generate scenes with more concrete object shapes and their detailed layouts. 


\paragraph{Indoor Scene Completion and Reconstruction.} A set of methods have been developed to complete or reconstruct 3D scenes from single \cite{song2017semantic,guo2018view,song2018im2pano3d,zhang2019cascaded} or multiple RGB and depth images \cite{dai2020sg}. Some object layout reconstruction methods \cite{nie2020total3dunderstanding, tulsiani2018factoring, popov2020corenet} recover the object pose, bounding box, and object layout from a single-view RGB image. All these methods are designed for reconstructing the geometry and semantic structure of a specific 3D scene and require a collection of complete 3D scenes for training. On the contrary, our method aims for generating different 3D scene layouts and is learned from a collection of semantic-segmented depth images. 

\paragraph{Learning 3D GAN from 2D Images.} A set of methods \cite{ProGAN2017,nguyen2019hologan,li2019synthesizing} have been proposed for learning the 3D GAN model of the objects in one category from 2D image collections. Different from these methods that focus on generating geometry or appearance of 3D objects from outside-looking-in images, our method focuses on generating indoor scene layout from inside-looking-out images. 

Although the volumetric GAN in our method is adapted from HoloGAN \cite{nguyen2019hologan}, these two approaches are different. By representing objects with feature volumes, The HoloGAN does not disentangle objects' shape and appearance and thus fails to generate a consistent projection of 3D objects under different views. Instead, our method models the scene geometry and layouts as semantic volume, which guarantees consistent projection from different views and is critical for learning 3D scene layouts from images captured from different unknown scenes. Also, different from HoloGAN that applies a single discriminator in training, we proposed a multiple-view discriminator for our task. 

\begin{figure*}[t]
    \centering
    \includegraphics{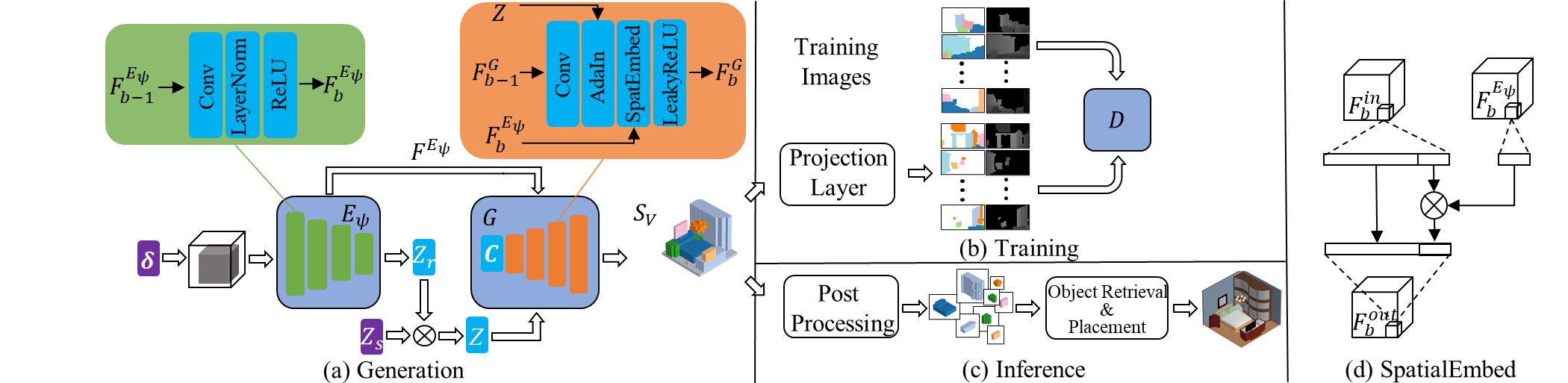}
    \caption{The overview of our method. (a) the volumetric generator;  (b) the projection layer and the discriminator that are used in the training stage; (c) the post-processing and object retrieval in the inference stage; (d) the detailed structure of SpatialEmbed used in our generator for fusing the conditional input of room size.}
    \label{FIG:Overview}
    \vspace{-0.2cm}
\end{figure*}

\section{Method Overview}
In this section, we provide an overview of the input, output, and basic components of our method. In the following sections, we discuss the technical details of each component of our method and the network training.  

\paragraph{Semantic-Segmented Depth Images.} The input of our method is a collection of semantic-segmented depth images ${I^{d,s}_i, i=1,2,\ldots,N}$ captured from different unknown rooms in a specific scene category, where the $I^d$ and $I^s$ refers to a depth image and its semantic label image. Each pixel in the semantic label image $I^{s}$ records the probabilities of the pixels' visible surface point belonging to each object class $c_j$ in ${c_0, c_1, c_j,\ldots,c_C, c_e}$, where $c_e$ is the label of empty space and $C$ is the number of all object classes in a specific scene category. For input semantic-segmented images, the semantic label vector in each pixel is a binary vector of $1$ for the ground truth object class of the pixel and $0$ for other object classes. For all input images, we assume that their camera's intrinsic parameters are known and each image includes a wall region of the scene so that we can estimate the camera's pose and distance for the visible part of the scene in each image. Since we have no information about the underlying 3D rooms of the input images, we have no idea whether two images are captured from the same room.

\paragraph{Volumetric Scene Representation.}
We represent a generated 3D indoor scene with a semantic scene volume $S_V$ with fixed spatial resolution $w\times{h}\times{d}$ ($32\times 32\times 16$ in our implementation), each voxel of which stores a probability vector of its semantic label ${p_0, p_1,\ldots, p_C, p_e}$. We align the floor of a scene with the XY plane of the volume and set the floor center to the center of the bottom volume layer $(h/2,w/2,0)$ ($(16,16,0)$ in our implementation). We predefine and fix the voxel's physical stride $\gamma$ for each scene category so that the maximal room size $(d\gamma,h\gamma,w\gamma)$ that can be modeled by the semantic scene volume is determined. Given a room with size $\psi=(R_x,R_y,R_z)$, we represent the layout, types, and shapes of the objects in the room with the semantic volume $S_V$, where all voxels out of the room boundary are labeled as empty.

\paragraph{System Overview.}
Our volumetric GAN consists of three main components: a generator $G$, a discriminator $D$, and a differentiable projection layer that connects the generator and the discriminator. As shown in Figure.~\ref{FIG:Overview}, the generator $G$ takes a latent vector $z_s$ and room size $\psi$ as input and outputs a semantic scene volume $S_V$ of the generated 3D scene. An encoder network $E_{\psi}$ encodes the room size $\psi$ into a set of conditional features of the generator. In the training stage, the projection layer renders the generated semantic scene volume $S_V$ from different views and feeds the rendered semantic-depth images to the discriminator $D$ to distinguish them from the real ones sampled from the training dataset. In the inference stage, we extract the object instances from the generated semantic scene volume in a post-processing step and then generate the final 3D scene by replacing all object instances with CAD models that are retrieved from an object database and best match the shapes and orientations of the object instances. 



\section{Network Design and Training}

\subsection{Generator}\label{SEC:ALG-GEN}
We adapt the volumetric network in HoloGAN\cite{nguyen2019hologan} as the basic network structure for our generator. Starting from a $2\times 2 \times 1$ constant feature volume with $512$ channels, our volumetric generator consists of four deconvolutional blocks used in \cite{nguyen2019hologan}, each of which reduces the number of feature channels by half and doubles the resolution of the feature volume along each dimension. We use the LeakyReLU as the activation function in the first three blocks and apply a softmax activation in the last block to output the probability of the semantic labels. As in \cite{nguyen2019hologan}, we use AdaIn to modulate the latent code $z$ via MLPs into the feature volumes after each block. 

To control the room size of the generated scene, we first generate a binary volume with the specified room size $\psi=(R_x,R_y,R_z)$, where all voxels within the room are labeled as $1$ and other voxels outside are marked as $0$. To control the generator with this binary volume, we encode the binary volume via a volumetric encoder $E_{\psi}$ with 4 convolutional blocks. The number of feature channels in each volumetric resolution is one-fourth of the ones of the volumetric generator layer with the same volume resolution and the length of the output feature $z_r$ is the same as the length of $z_s$. We modulate latent vector $z_s$ with the room control feature $z_r$ via a dot product $z=z_s\cdot{z_r}$. Meanwhile, we modulate $1/4$ channels of each feature volume in the generator with the feature volume of $E_{\psi}$ in the same volume resolution via element-wise dot products, as shown in FIG.~\ref{FIG:Overview}(d). With these two modulations, our method can successfully constrain the scene generation within the room volume with the specified size. 



\subsection{Differentiable Projection Layer}\label{SEC:ALG-PROJ}
Given a viewpoint, we apply the differentiable ray consistency (DRC) \cite{tulsiani2017multi} to render the depth and semantic images from the generated semantic volume $S_V$. Specifically, we take the probability of the "empty" label as the non-occupied probability of scene voxels and exactly follow the DRC for computing pixel depth. To render the probability vector of the semantic labels in each pixel, we regard the probability of each object category as an independent voxel property and compute its pixel value via DRC. After that, we concatenate the values of all object categories and the accumulated probability of the "empty" label to get the probability vector of semantic labels for each pixel. 

To make sure that the rendered images follow the same view distributions of the training images, we render the images from semantic scene volume with the same camera settings (i.e. intrinsic parameters, distance and pose to the visible room wall) as the ones of the training images. 



\begin{figure}
    \centering
    \includegraphics{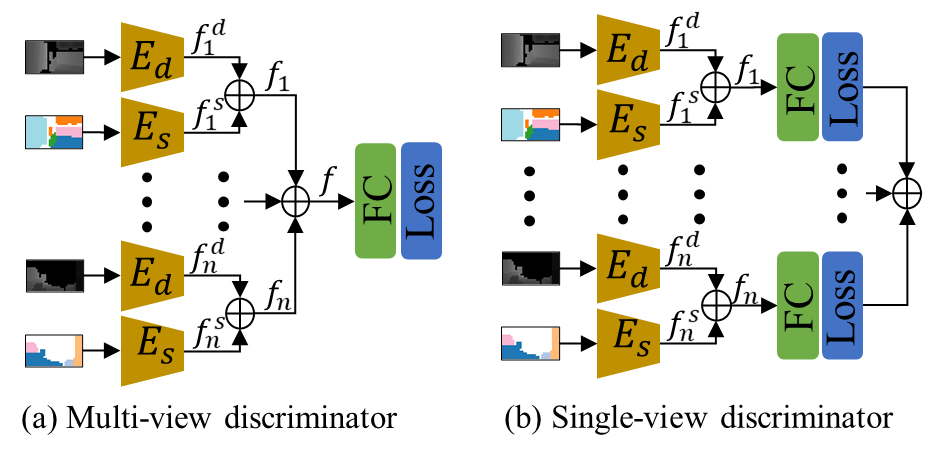}
    \caption{The design of our multiple-view discriminator. (a) the network structure of our design; (b) A naive sum of multiple single-view discriminators. }
    \label{FIG:Discriminator}
\end{figure}

\begin{figure}
    \centering
    \includegraphics[width=0.45\textwidth]{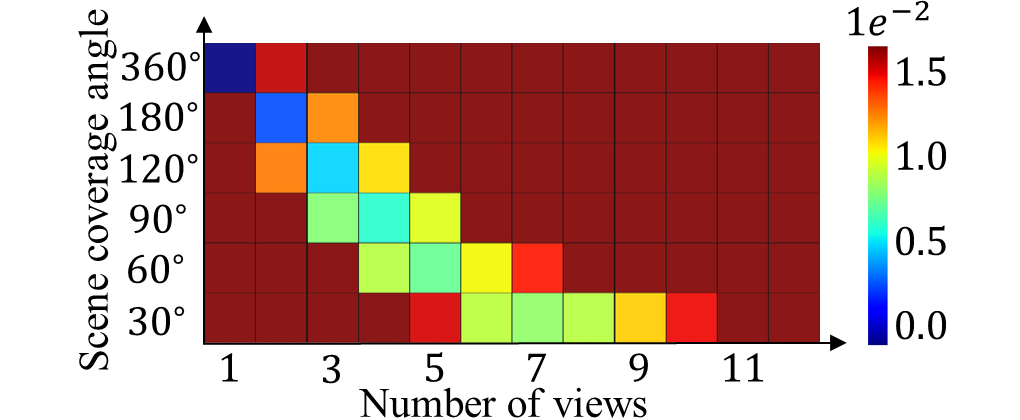}
    \caption{A visualization of the differences between the object co-occurrence maps of the ground truth scenes and the ones of random view combinations under different numbers of views (columns) and scene coverage angles (rows). The dark red indicates larger difference caused by a poor approximation of the scene layouts, while the blue refers to the small difference of a good approximation. Any differences larger than $1.5e^{-2}$ are caused by poor approximations and mapped to $1.5e^{-2}$ in this visualization.}
    \label{FIG:ViewAngle}
\end{figure}

\subsection{Multi-View Discriminator Network}\label{SEC:ALG-DIS}
Our discriminator $D$ takes $N$ semantic-segmented images $I^{d,s}$ as input and computes a score for the loss function. 

\paragraph{Multi-View Discriminator} A naive design of the discriminator is to apply the single-view discriminator in \cite{nguyen2019hologan} or a sum of several single-view discriminators as in \cite{li2019synthesizing} for our task. Unfortunately, we find that this design is prone to generating poor results. The possible reason is that scene generation needs to address larger depth variation than object generation, the differentiable ray consistency (DRC) tends to drive the generator to create scenes with objects closer to the viewpoint with the single-view discriminator. Although the generated scene is different from the GT, their single view rendering still matches some images in the training dataset and thus passes the discriminator. Please refer to the supplemental for more detailed discussions. 

To solve this problem, we design a multi-view discriminator, where we extract features of multiple views captured from a scene and then compute a loss function of jointed features. Given multiple semantic-segmented images, we use two feature extractors $E_d$ and $E_s$ that share the same network structure but independent weights to encode the depth and semantic images respectively. Each network consists of four convolutional blocks, each of which includes a convolution layer to reduce the image resolution to half and double the number of feature maps, a spectrum norm layer, and a ReLU layer. The reshape layer outputs a 512-length feature vector. After that, we concatenate the feature vectors of all input images and feed them into the last FC layers to get the final score. Fig.~\ref{FIG:Discriminator} illustrates the difference between our multi-view discriminator (Fig.~\ref{FIG:Discriminator}(a)) and a naive combination of single-view discriminators (Fig.~\ref{FIG:Discriminator}(b)).

\paragraph{View Configuration for Discriminator} 
For generated 3D scenes, we can sample multiple views from one scene to cover the scene so that the multi-view discriminator can get the complete scene layout information in the training. However, we cannot obtain the layout of the underlying ground truth scenes in the training dataset since we have no scene information for each image.


To solve this issue, we use a random combination of the training images to approximate the layouts of the underlying ground truth scenes. To this end, we seek an optimal view configuration (i.e. the number of views and the coverage of each view) of image combinations that can best approximate the ground truth scene layouts. For this purpose, we render the panorama images of a set of 3D rooms modeled by the artist from the room center and then split the panorama images into the different number of views, each of which corresponds to a view coverage setting. For each specific view coverage, we generate a set of image combinations, each of which includes a specific number of images randomly picked from the view collection. After that, we compute the difference between the co-occurrence map\cite{li2019grains} of the objects in the ground truth 3D rooms and the co-occurrence map of the objects in the image combinations with a specific number of views and view coverage. Because the co-occurrence map provides the first-order statistics of the object distributions in the scene, it provides a reasonable indication of how well the image combinations can approximate the object distributions of the underlying scenes. Fig.~\ref{FIG:ViewAngle} shows the heat map of the differences for different view number and coverage configurations, where the view configurations in dark red result in a poor approximation of the underlying scene layouts, and the blue and green ones offer a relatively good approximation. Based on this empirical analysis, our method sets the view angle of each image around $110^\circ$ and includes 4 images randomly picked from the training dataset in one image combination for network training. 

\begin{figure}
    \centering
    \includegraphics{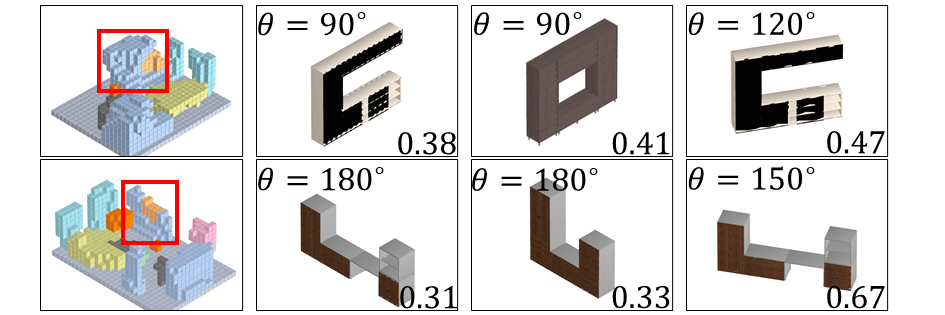}
    \caption{An illustration of the object retrieval step. Given an instance of the TV stand ((highlighted with red boxes in the first column), our method finds three object candidates from the database and their rotation angles (in the following columns) that best match the shape and orientation of the input. Note that for two instances of TV stands that have  similar bounding boxes but different overall shapes shown in the different rows, our method can find different CAD models for each instance.}
    \label{FIG:Retrieval}
    \vspace{-0.3cm}
\end{figure}

\subsection{Network Training}
\paragraph{Loss Function}
We follow the training scheme in \cite{nguyen2019hologan} to train our volumetric GAN with the new loss functions defined for our task. Specifically, the loss for the generator is defined as 
\begin{align}
    L_G(Z_S, \Psi)=&\sum_{z_s\sim{Z_S},\psi\sim{\Psi}}log(D(P(G(z_s,\psi)))\nonumber\\
    +&(-M_0(\psi)log(G(z_s,\psi))),
\end{align}
where the first term is the generative loss and the second term is the cross-entropy loss for constraining the voxels that are out of the room boundary in the generated scene volume to be "empty". Here $P$ is the differentiable projection layer and $M_0$ is the empty voxels in the mask volume of $\psi$.

The loss for the discriminator is defined as:
\begin{align}
    L_D(Y, Z, \Psi)=&\sum_{z\sim{Z},\psi\sim\Psi}(1-log(D(P(G(z,\psi))))\nonumber\\
    +&\sum_{y\sim{Y}}log(D(y))
\end{align}
where $y\sim{Y}$ is the set of real joint-views images selected from the training dataset.

\begin{figure}
    \centering
    \includegraphics[width=0.45\textwidth]{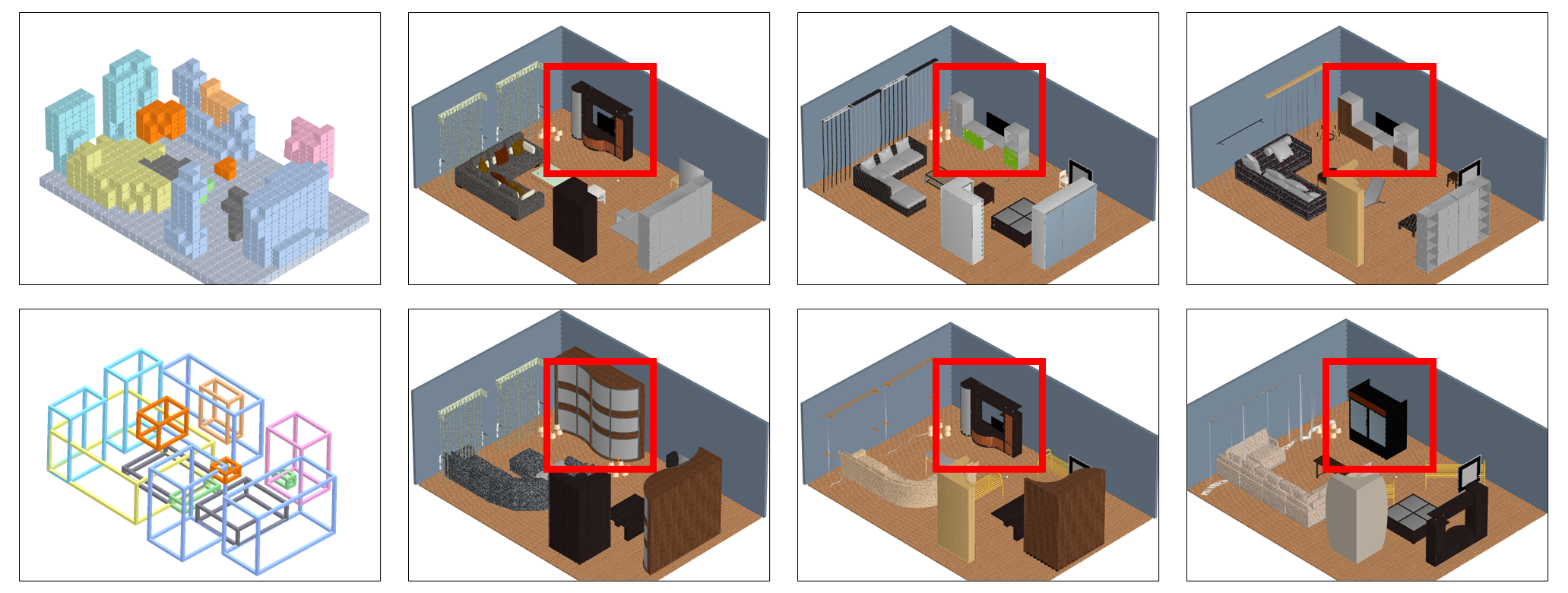}
    \caption{The impact of the object instance volume to object retrieval. Given an input semantic scene volume generated by our method (shown in the first column), we compare the scene generated by retrieving the CAD models with the extracted object instance volumes (the first row) and the one generated by retrieving the bounding boxes of the extracted object instances (the second row). Note that our method can successfully generate the detailed object layouts of a scene (e.g. the TV and TV stand in the red box), while the bounding-box based method fails.}
    \label{FIG:RetrievalUsingDetails}
    \vspace{-0.3cm}
\end{figure}

\section{3D Indoor Scene Generation}\label{SEC:ALG-PPR}
Given a room size $\psi$ and a latent vector $z_s$ randomly sampled from the latent space, the trained volumetric GAN can generate a semantic scene volume that stores both layout and rough shapes of the objects instances in the room. To generate the final 3D scene, we extract the object instances from the semantic scene volume and replace them with the CAD models retrieved from a 3D object database. 

\paragraph{Post-Processing} To acquire the object instances from the semantic scene volume, we first set the semantic label of each voxel as the one with the maximal probability. We randomly pick a voxel and iteratively group its neighboring voxels with the same label as an object instance via the flood filling algorithm. After that, we mark all voxels in the object instance is processed. We repeat this process until all voxels in the volume are processed. To remove the outliers, we discard the object instances the sizes of which are smaller than the minimal size of objects in the same object class of the 3D object database. Our experiments show that only around $1\%$ of object instances are removed in this step. 


\paragraph{Object Retrieval and Placement} For each object instance $\hat{M}$ extracted from the semantic scene volume, we search the 3D object  database $E(M)$ to find a 3D object $M_i$ and its rotation $\phi$ along $Z$ axis so that $M_i$ and $\hat{M}$ belong to the same object class and the rotated $M_i^{\theta}$ best matches the shape of the $\hat{M}$ and has minimal collision with the surrounding objects in the volume:
\begin{align}
    i^*,\phi^*=\argmin_{i\in{I},\phi\in\Phi}{CD}(\hat{M},M_i^\phi)+\lambda{w_c}({M_i^\phi})
\end{align}
where $CD$ is the Chamfer distance between the candidate instance $\hat{M}$ and the rotated object $M_i^\phi$. $w_c$ is the penalty term of spatial collision, defined by the IoU between the rotated model $M_i^\phi$ placed in the scene and the surrounding voxels with other object class ID. $\lambda$ is a scalar ($1.0$ in our implementation) to balance the distance and collision terms.

Figure.~\ref{FIG:Retrieval} illustrates the results generated by our object retrieval and placement algorithm. Note that the volumetric representation offers the rough shape and orientation of the objects in the scene and thus result in detailed object layouts (e.g. TV and TV stands) that are difficult to be modeled by the object's bounding box. 
\section{Experimental Results}\label{SEC:RES}

\begin{table}[t]
    \centering
    \scalebox{0.85}{
    \begin{tabular}{c|ccc}
        \hline
        Dataset                 & images & scenes & object classes   \\ \hline
        Structured3D-Bedroom    & 16064  & 5219   & 9            \\ \hline
        Structured3D-Livingroom & 6592   & 2211   & 11           \\ \hline
        Structured3D-Kitchen    & 3009   & 1491   & 5            \\ \hline
        Matterport3D-Bedroom    & 1217   & 178    & 8            \\ \hline
        NYUv2-RGB-Bedroom       & 1495   & 119    & 10           \\ \hline
    \end{tabular}
    }
    \caption{The number of scenes, images, and object classes in each training dataset used in our experiments. }
    \vspace{-0.3cm}
    \label{TAB:scene}
\end{table}

\paragraph{Implementation Details} We implement our algorithm with Tensorflow and train our GAN model on a machine with 4 TESLA V100 GPUs. We train the network via the Adam optimizer. The sizes of semantic scene volumes and images are $32\times 32\times 16$ and $32\times 18$ respectively. The learning rate is $2e^{-4}$ and the batch size is 128. The training converges after 2,000 epochs. 

\paragraph{Training Dataset} We test our method on semantic-segmented depth images datasets in Structured3D \cite{zhang2020deep} and Matterport3D \cite{chang2017matterport3d} scene collections. We also apply our method to semantic-depth images inferred RGB images of NYUv2 \cite{silberman2012indoor} dataset. In all experiments, we use the ShapeNet dataset\cite{wu20153d} for retrieving and placing CAD models to generate final 3D scenes.  


For Structured3D, we train our model on three scene categories (Bedroom, Living room, and Kitchen) that exhibit rich layout variations. For each scene category, we collect the image by checking the angles between the ray to the room center and the optical axis of the images and choosing all images with angles smaller than $\pm{45}^\circ$. After that, we downsample the images to $32\times18$. For this purpose, we first project the semantic-segmented depth images back into the 3D space and then voxelize the 3D space into a $32\times32\times16$ volume. Finally, we render the volume to the $32\times18$ semantic-segmented depth images from the original viewpoint. For each scene category, we also merge the classes of objects with similar functionalities into one and remove the classes of objects that appear infrequently in scenes. The class ID of the pixels rendered from the removed objects is set to empty. For Matterport3D, we follow the same procedure to collect images from the bedroom category in our experiment.  

For the NYUv2 dataset, we generate depth and semantic-segmented maps for all unlabeled RGB images in the bedroom scene category with the method in \cite{nekrasov2019real}. We then manually remove noisy results and select 1459 images with relatively large view coverage. After that, we follow the procedure described above to generate the training image set from the inferred semantic-segmented depth images. The statistics of the training dataset used in our experiments are listed in Table.~\ref{TAB:scene}.

\begin{figure}
    \centering
    \includegraphics{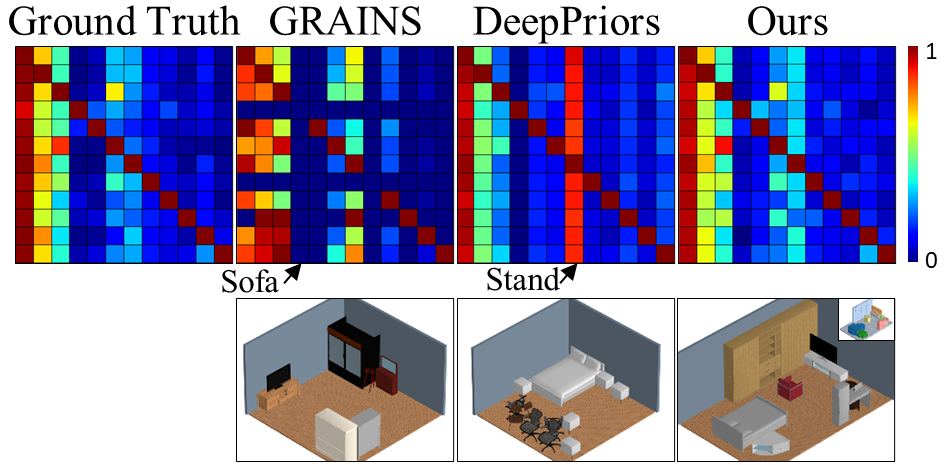}
    \caption{The co-occurrence maps of the ground truth scenes and the ones generated by GRAINS\cite{li2019grains}, DeepPrior\cite{wang2018deep}, and our methods. Note that the co-occurrence map of our method is more similar to the ground truth one than the ones generated by GRAINS and DeepPrior, especially for values in the columns indicated by the arrows. The second row illustrates the typical unnatural scenes generated by GRAINS and DeepPrior, as well as a 3D scene generated by our method. Please refer to the supplemental material for all co-occurrence maps of scenes generated by different methods.}
    \label{FIG:CmpPriors}
    \vspace{-0.3cm}
\end{figure}

\subsection{Method Evaluation}\label{SEC:Res-Compare}
To validate our network design and the advantages of our method to other existing solutions, we train our model with the semantic-segmented depth images rendered from the same scene category of the ground truth dataset in \cite{li2019grains} for a fair comparison. 

We first compare the co-occurrence map \cite{li2019grains} of the objects in scenes generated by our method and the pre-trained models in DeepPrior\cite{wang2018deep} and GRAINS\cite{li2019grains}. For any two object classes, their co-occurrence map value is the ratio between the number of scenes in which two object classes appear together to the number of scenes that only one object class (shown in the row) appears.
As shown in Fig.~\ref{FIG:CmpPriors}, the co-occurrence map of our method is much more consistent with the ground truth than the ones of DeepPrior and GRAINS. In particular, GRAINS\cite{li2019grains} generates many fewer sofas in the resulting scenes than the ground truth, while DeepPrior\cite{wang2018deep} generates many more stands in the results (the columns indicated by the arrow in Fig.~\ref{FIG:CmpPriors}). 

\begin{figure}
    \centering
    \scalebox{0.9}{
    \includegraphics[width=0.45\textwidth]{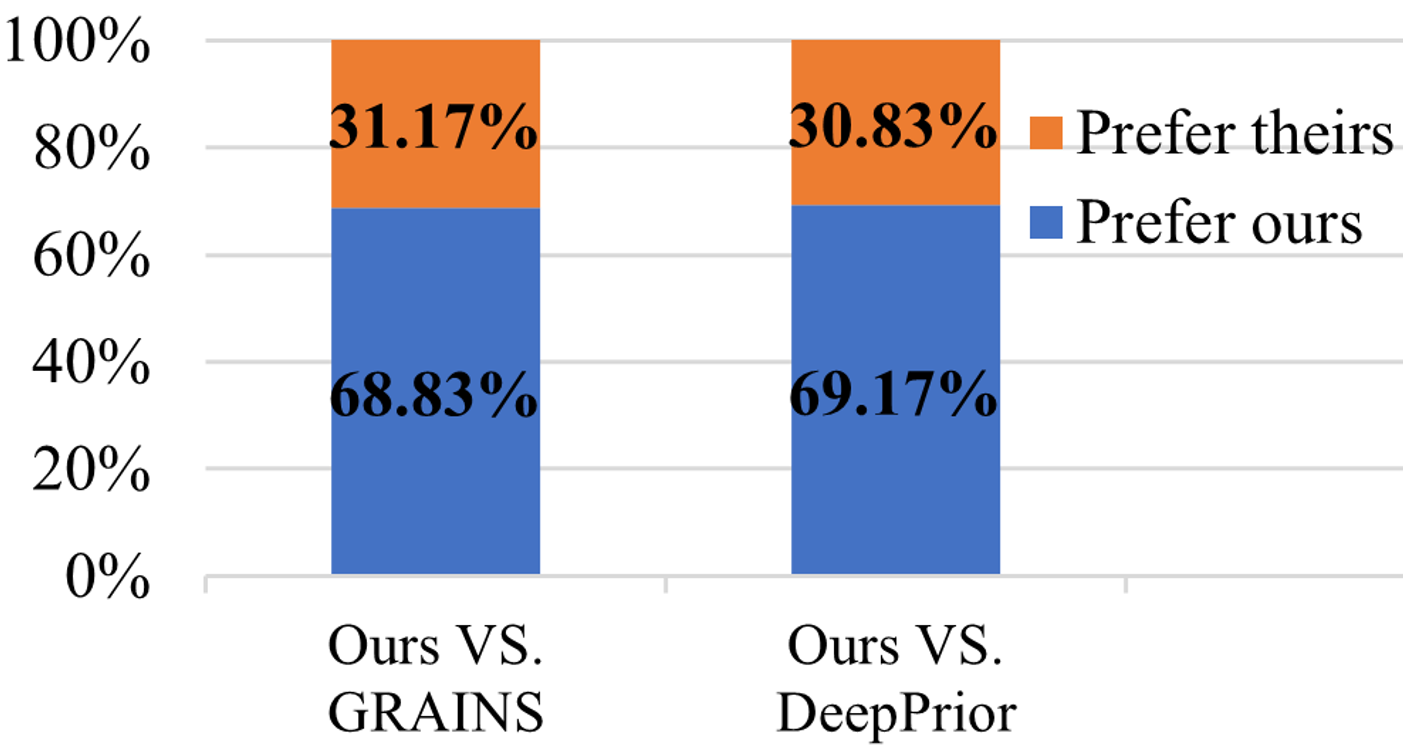}
    }
    \caption{The percentages of the methods that are selected by the participants in our user study. In both experiments, the participants prefer the results generated by our method much more than the ones generated by GRAINS\cite{li2019grains} and DeepPrior\cite{wang2018deep}}
    \vspace{-0.2cm}
    \label{FIG:UserStudy}
\end{figure}

We further conduct two user studies for comparing the visual quality of scenes generated by our method with the ones generated by GRAINS and DeepPrior separately. In each study, we show the images rendered from three scenes with the same rendering settings: one generated by our method, one generated by the existing method, and the ground truth reference with similar object classes to the generated ones. We show the same set of $30$ groups of images to $20$ participants. For each group of three images, we ask each participant to choose one from two generated results that are more plausible to the reference. For GRAINS and DeepPrior, we only render the images of the scenes that match the co-occurrence map of the ground truth in our user study for a fair comparison. As shown in Fig.~\ref{FIG:UserStudy}, the users prefer scenes generated by our method more than the ones generated by other existing methods. 

Finally, we evaluate the diversity and quality of 3D scenes generated by our method. For diversity, we implement the similarity metric in \cite{wang2019planit} and the average similarity of the generated scene layouts and GT are $0.335$(Gen) and $0.457$(GT) respectively, which indicates that our method well preserves the variation of the scene layouts in the training dataset. For quality, we follow the method in \cite{ritchie2019fast} and evaluate the real vs. synthesis classification accuracy of our method by training a classifier with $800$ semantic-segmented depth images of the generated scenes rendered from random viewpoints and 800 GT semantic-segmented depth images. After that, we compute the accuracy score with $320$ semantic-segmented depth images rendered from another set of generated scenes and the result is $60.9\%$, which illustrates that our results are difficult to be classified as real or fake.

\begin{table}[t]
    \centering
    \scalebox{0.8}{
    \begin{tabular}{c|cccc}
        \hline
        Methods     & SD-MV-D  & USD-MV-D & SV-D     &Ours    \\ \hline
        Acc. Ratio  & 64.0     & 69.0     & 71.0     &80.0    \\ \hline
        Methods     & 2View-MV-D& 6View-MV-D& 8View-MV-D      \\ \hline
        Acc. Ratio  & 70.7     & 72.0     & 70.3              \\ \hline
    \end{tabular}
    }
    \caption{The results of ablation study, where the numbers are acceptance ratios of the scenes generated by different network configurations.}
    \label{TAB:Ablation}
    \vspace{-0.3cm}
\end{table}

\begin{figure*}
    \centering
    \scalebox{0.9}
    {
    \includegraphics{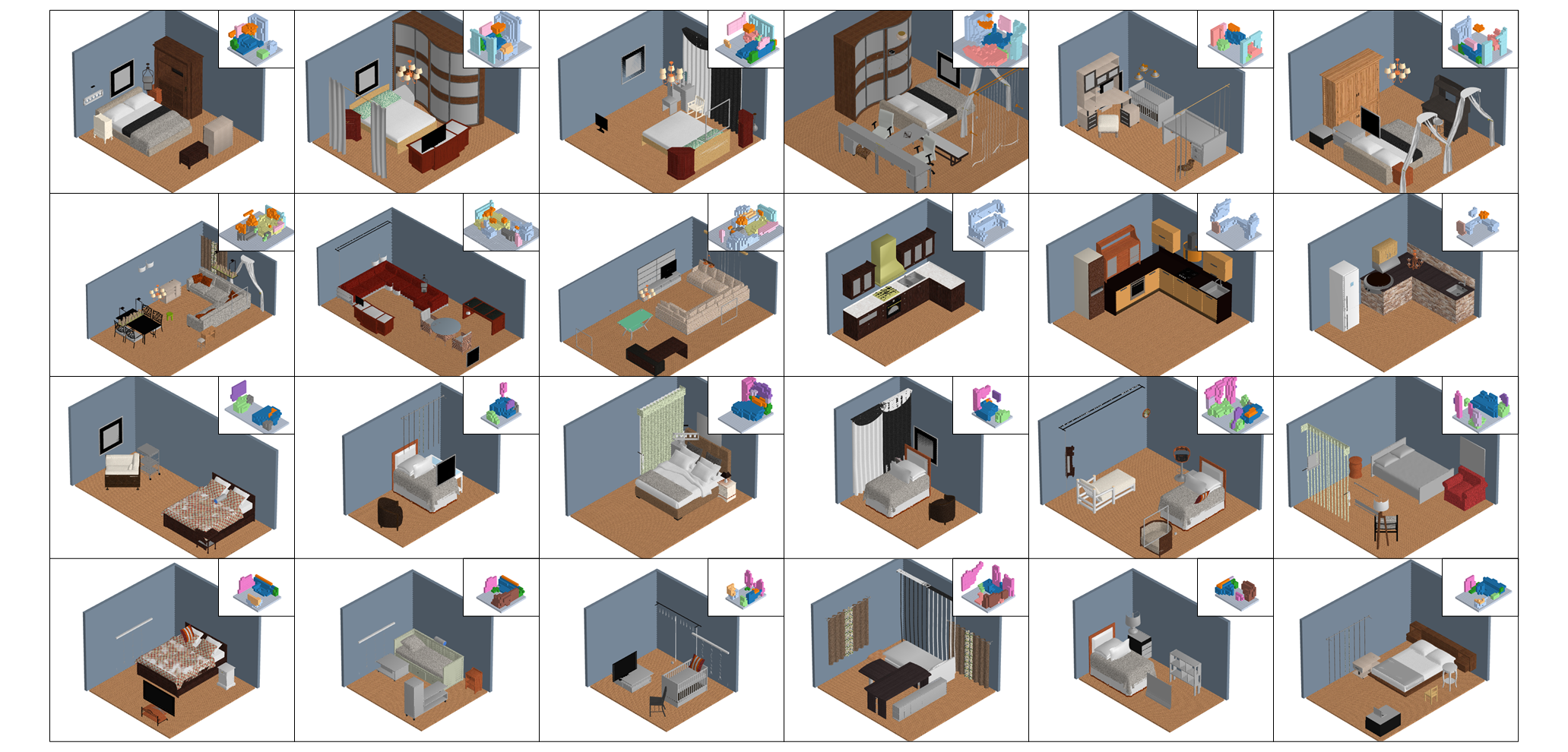}
    }
    \caption{The 3D scene generated by our method for different scene categories. The first row shows the results generated from Structured3D-Bedroom dataset. The left three results in the second row are the results generated from the Structured3D-Livingroom dataset  and the right three results are results generated from the Structured3D-Kitchen dataset. The third and last row illustrate the results generated from the Matterport3D-Bedroom and NYUv2-RGB-Bedroom datasets, respectively. The semantic scene volume of each result is shown at top-right corner of each image.}
    \label{FIG:Visual}
    \vspace{-0.2cm}
\end{figure*}

\subsection{Ablation Study}\label{SEC:Res-Valid}
We conduct a set of ablation studies on Structured3D-Bedroom dataset to validate our network design and list all results in Table.~\ref{TAB:Ablation}, where "Ours" is the result of our current network design. 

Due to the statistical similarity of the results in different network configurations, we find the co-occurrence map and user study metric used in Sec.~\ref{SEC:Res-Compare} cannot clearly indicate the difference of different network configurations. Instead, we ask $5$ experienced users who can identify the failure cases based on their prior knowledge to pick failure cases from $50$ scenes randomly generated by each network setup and then compute the acceptance ratio of the result for comparison.

\vspace{-0.3cm}
\paragraph{Single-View  vs. Multi-View Discriminator} To validate the advantage of our multi-view discriminator, we train the GAN model with a single-view discriminator shown in Fig.\ref{FIG:Overview}(c). Compared to multi-view discriminator (Ours), the acceptance ratio of the GAN trained with single-view discriminator (SV-D) decreases from $80.0$ to $71.0$. 

\vspace{-0.3cm}
\paragraph{Unified vs. Separated Encoder for Semantic/Depth} Our method uses a different encoder for extracting features from semantic and depth channels in a multi-view discriminator. An alternative solution is to stack the semantic and depth channels together and extract features with one unified encoder. Our method outperforms the model to this alternative encoder scheme (USD-MV-D) with a $10.7$ acceptance ratio gap ($80.0$ VS. $69.3$). 

\vspace{-0.3cm}
\paragraph{Unified vs. Separated Multi-View Discriminator}
Instead of one unified multi-view discriminator for both depth and semantic, another design uses a separate multi-view discriminator for semantic images and depth images separately. This design (SD-MV-D) is worse than our current scheme with $16.0$ acceptance ratio gap ($64.0$ vs. $80.0$). 

\vspace{-0.3cm}
\paragraph{Number of Views in Multi-View Discriminator} Our current network applied a multi-view discriminator with 4 views for GAN training. To analyze the performance of our model with different numbers of views used in multi-view discriminators, we change the number of view branches in the multi-view discriminator to 2, 6, and 8 respectively, and train the GAN model with the same image set used in our model training. As shown in Table.~\ref{TAB:Ablation}, our method achieves the best performance, while models with fewer views (2view-D) or more views (6view-D and 8view-D) generate worse results. 

\vspace{-0.3cm}
\paragraph{Volumetric Representation and Resolution}
The semantic volume representation successfully models the rough object shapes and their detailed layout. As shown in Figure \ref{FIG:Retrieval} and Figure. \ref{FIG:RetrievalUsingDetails}, it provides better input for object retrieval and placement than the object's bounding box and generates a detailed layout of the objects that are difficult to be modeled by existing methods. The low-resolution semantic volume $32\times32\times16$ used in our current implementation achieves a good balance between model capability and computational cost. We train a GAN model for generating $64\times64\times32$ volume and the results are similar. 

\subsection{Visual Results}
Fig.~\ref{FIG:Visual} visualizes the 3D scenes generated by our method from different scene categories and datasets, including the Structured3D-Bedroom (1st row), Structured3D-Livingroom (the first three in 2nd row), Structured3D-Kitchen (the last three in 2nd row), MatterPort3D-Bedroom (3rd row), and NYUv2-RGB-Bedroom (4th row). For all these scene categories, our GAN model successfully learns and generates various scene layouts for both large (e.g. cabinets and beds) and small objects (e.g. ceiling lamps, pictures). Also, our method is robust to both synthetic scenes and real scenes, as well as the semantic-segmented depth images inferred from the real RGB images. Thanks to the semantic scene volume representation, our method can generate non-cubic Manhattan layouts and detailed scene layouts, such as chairs with different orientations, a chair under a desk, and a TV inside a TV cabinet. More visual results can be found in the supplemental.



\section{Conclusion}
We propose a GAN model learned from semantic-segmented depth images for 3D scene generation. To this end, we model the scene layout with semantic scene volume and propose a new multiple-view discriminator for efficient GAN training. Our method greatly reduces the workload for capturing or modeling 3D scenes and generates good results, including the detailed scene layouts that are difficult to be model using previous approaches. 

Our method still has some limitations. Since the semantic scene volume does not have furniture instance information, our method requires post-processing to separate furniture instances from the volume for object retrieval. It is interesting to explore a new scene layout representation that can well model furniture instances in a scene. Also, our method still needs post-processing and object retrieval for generating the resulting 3D scene. A promising direction in this area is to develop a new method that can directly synthesize the detailed 3D scene with different scene layouts. Finally, it is interesting to learn a generative model of 3D scenes from a collection of RGB images. 

\paragraph{Acknowledgements}
We would like to thank anonymous reviewers for their constructive comments and suggestions, and Yue Dong and Nathan Holdstein for their help to proofread the paper.

{\small
\bibliographystyle{ieee_fullname}
\bibliography{egbib}
}

\clearpage



\section*{Supplementary material}

# Indoor Scene Generation from a Collection of Semantic-Segmented Depth Images – Supplemental Material

Ming-Jia Yang[*,1,2]   Yu-Xiao Guo[2]   Bin Zhou[1]   Xin Tong[2]
[1]Beihang University    [2]Microsoft Research Asia
{yangmingjia,zhoubin}@buaa.edu.cn   {yuxgu,xtong}@microsoft.com

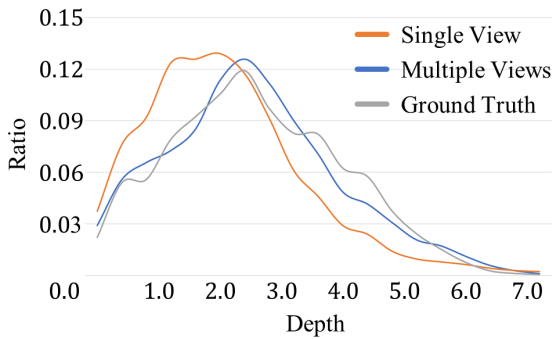

Figure 1. The distribution of the depth of all depth images of the scenes rendered from the ground truth scenes (grey curve) and the scenes generated by GAN with single-view discriminator (orange curve) and our multi-view discriminator (blue curve).

## 1. The Failure of Single-View Discriminator in 3D Scene Generation

We conduct a preliminary experiment and analysis to investigate the reason behind the failure of the single-view discriminator [3, 2] in our scene generation task.

To this end, we train our volumetric GAN model with the Structured3D-Bedroom dataset. For each scene in the dataset, we set the camera at the room center and 1.5m height to the floor and render four depth images by aligning the camera view to the $+X, -X, +Y, -Y$ directions. We then take these rendered images as training data for learning our volumetric GAN model with a single-view discriminator and our multi-view discriminator separately. After that, we generate 5,000 scenes for each network and render the depth images with the same camera setup as the training images. For depth images of the ground truth scenes and the scenes generated by each network, we plot the distributions of the depth values of all images in Fig. 1. We find that different from the 3D object modeling task in [3, 2] where the objects in the images have similar distances to the viewpoints of images, the objects can be anywhere in a room so

| Dataset | Selected categories |
|---|---|
| Structured3D-Bedroom | Cabinet, Bed, Chair, Picture, Desk, Curtain, Television, Night stand, Lamp |
| Structured3D-Livingroom | Cabinet, Chair, Sofa, Table, Picture, Shelves, Curtain, Lamp, Pillow, Refrigerator, Television |
| Structured3D-Kitchen | Cabinet, Picture, Curtain, Refrigerator, Lamp, |
| Matterport3D-Bedroom | Bed, Pillow, Night stand, Chair, Picture, Lamp, Curtain, Table |
| NYUv2-RGB-Bedroom | Cabinet, Bed, Chair, Desk, Shelves, Curtain, Pillow, Television, Nigh Stand, Lamp |

Table 1. The object classes in each training dataset.

that the distances between the viewpoint and objects in the scene have larger variations than the view distances between the camera and 3D objects. For images with such large depth variations, the differentiable ray consistency (DRC) layer used in our network tends to drive the generator to create scenes with more objects closer to the viewpoint with the single-view discriminator (the orange curve in Fig. 1). Although the generated scene is different from the GT, their rendering matches some views in the training dataset and thus could pass the discriminator and results in the model collapse. With the joint-views used in our multi-view discriminator, the model collapse caused by the single-view discriminator could be avoided and the depth distribution of the generated scene (the blue curve in Fig. 1) matches the depth distribution of the GT scenes (the grey curve in Fig. 1) better. One possible reason is that among all random combination of training images, the percentage of combinations in which all images in the combination are close-up views becomes much fewer.

## 2. More Experimental Results

**Details of Object Classes**   Table. 1 lists the list of the object classes in each training dataset.

**The Complete Co-Occurrence Maps**   Fig. 2 illustrates the object co-occurrence maps of the GT scenes, the scenes

---
[*]This work is done when Ming-Jia Yang was an intern at MSRA

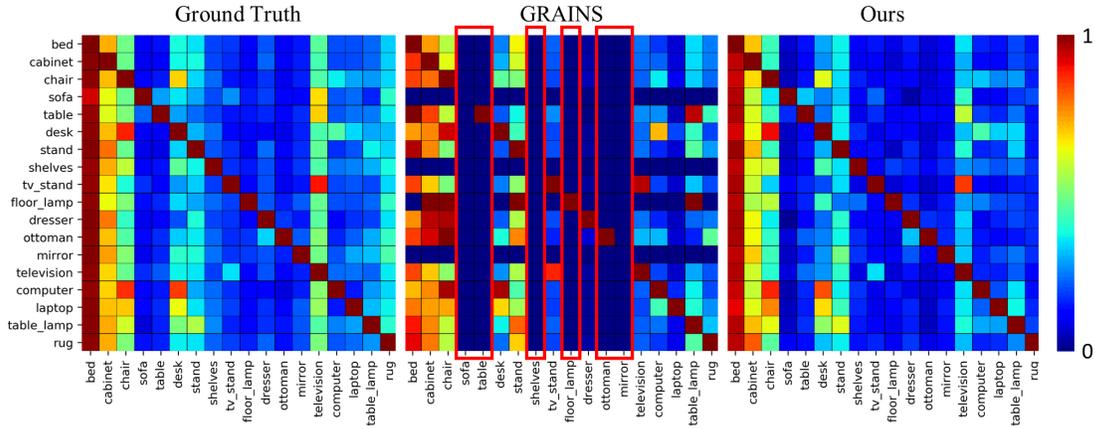

Figure 2. The object co-occurrence maps of all object classes of the ground truth scenes, scenes generated by GRAINS[1], and scenes generated by our method.

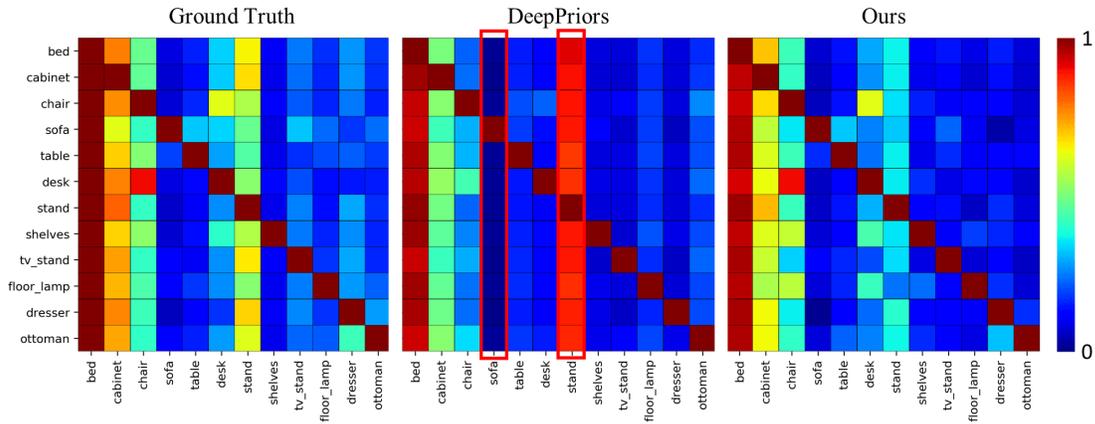

Figure 3. The object co-occurrence maps of all object classes of the ground truth scenes, scenes generated by GRAINS[1], and scenes generated by our method.

generated by our method, and the scenes generated by GRAINS [1] for all object classes in the scene. Fig. 3 visualizes the object co-occurrence maps of the GT scenes, the scenes generated by DeepPrior[4], as well as scenes generated by our method for all object classes in the scene. These two figures are a complete version of the co-occurrence maps shown in the paper.

**The User Interface**   Fig. 4 and Fig. 5 displays user interface used in our user study and the user interface used in our ablation study.

**More Visual Results**   Here we show more images of 3D scenes generated by our method from different training datasets, including Structured3D-Bedroom (Fig. 6), Structured3D-Livingroom and Structured3D-Kitchen (Fig. 7), Matterport3D-Bedroom and NYUv2-RGB-Bedroom (Fig. 8).

## References

[1] Manyi Li, Akshay Gadi Patil, Kai Xu, Siddhartha Chaudhuri, Owais Khan, Ariel Shamir, Changhe Tu, Baoquan Chen, Daniel Cohen-Or, and Hao Zhang. Grains: Generative recursive autoencoders for indoor scenes. *ACM Trans. Graph.*, 38(2):1–16, 2019. 2

[2] Xiao Li, Yue Dong, Pieter Peers, and Xin Tong. Synthesizing 3d shapes from silhouette image collections using multi-projection generative adversarial networks. In *IEEE Conf. Comput. Vis. Pattern Recog.*, pages 5535–5544, 2019. 1

[3] Thu Nguyen-Phuoc, Chuan Li, Lucas Theis, Christian Richardt, and Yong-Liang Yang. Hologan: Unsupervised learning of 3d representations from natural images. In *Int. Conf. Comput. Vis.*, pages 7588–7597, 2019. 1

[4] Kai Wang, Manolis Savva, Angel X Chang, and Daniel Ritchie. Deep convolutional priors for indoor scene synthesis. *ACM Trans. Graph.*, 37(4):1–14, 2018. 2

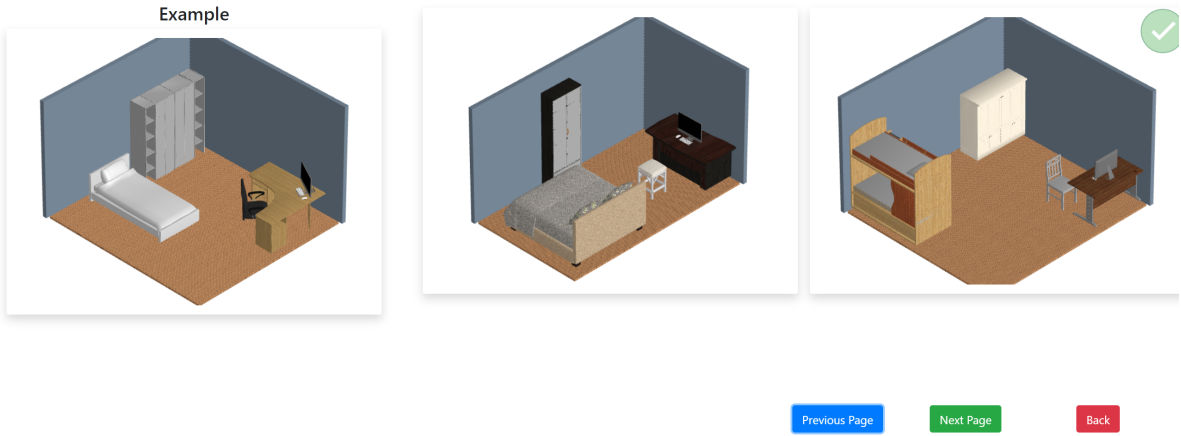

Figure 4. The user interface of our program used in the user study.

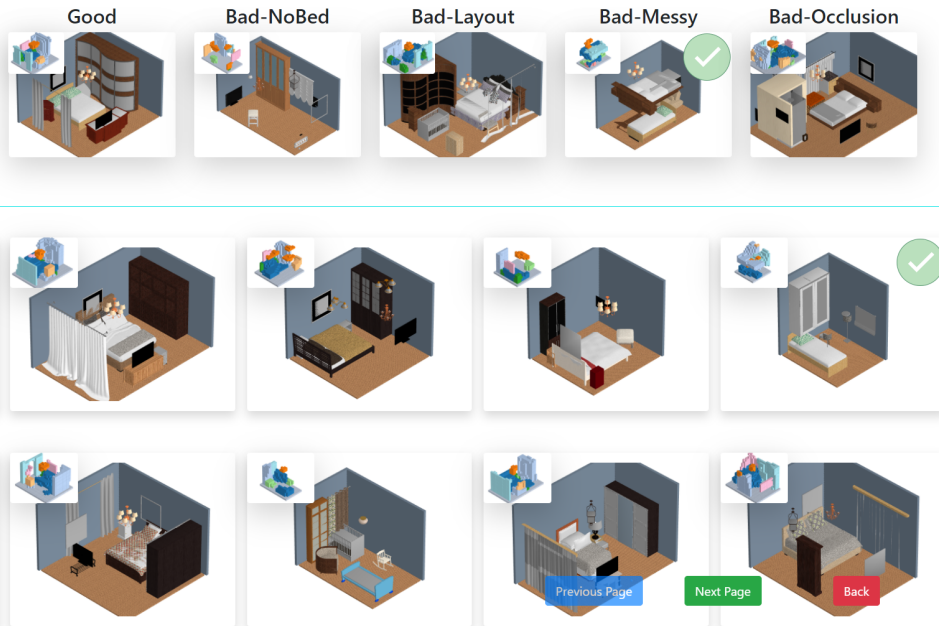

Figure 5. The user interface of our program used in the ablation study.

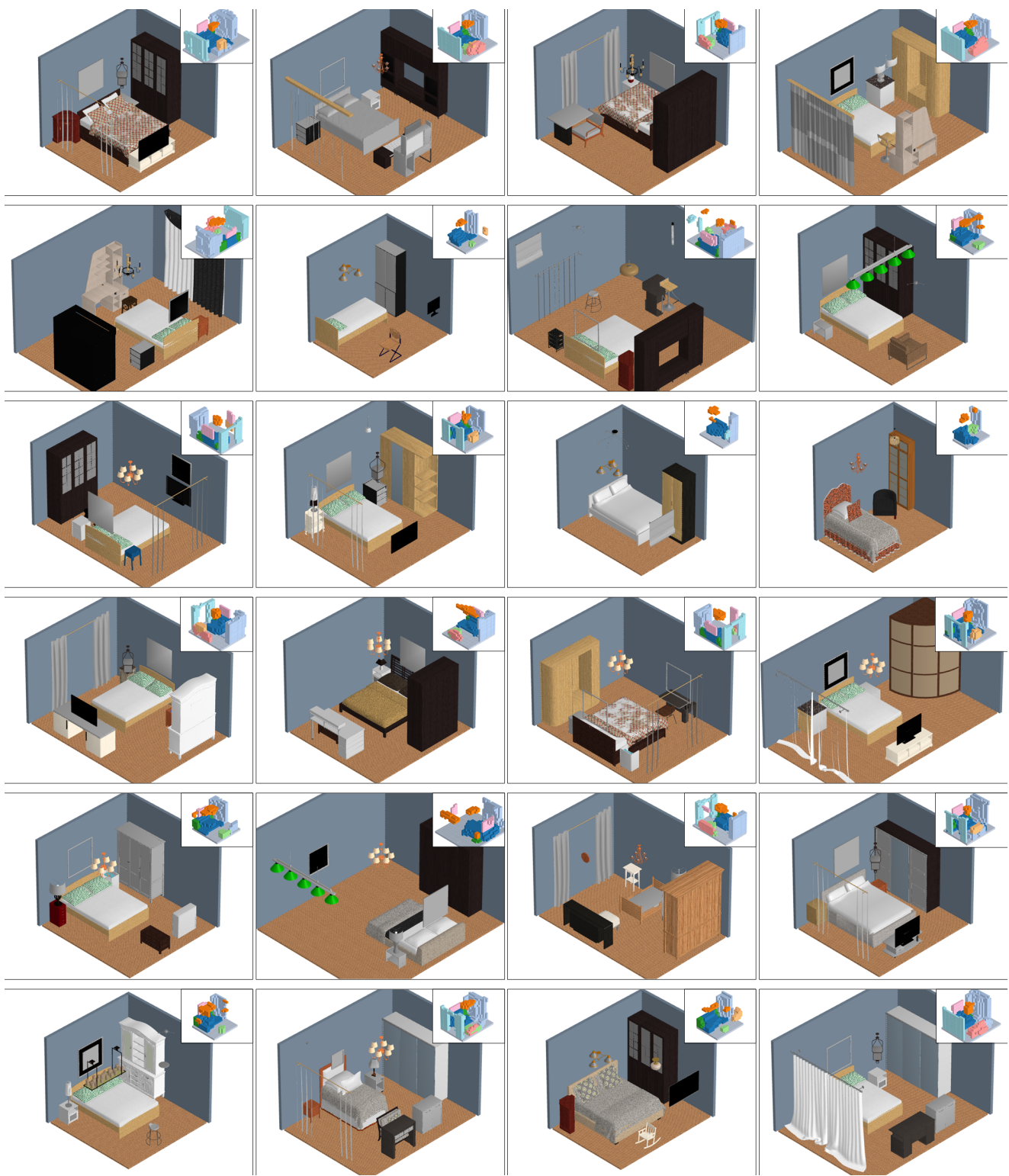

Figure 6. More results generated by our method from the Structured3D-Bedroom dataset.

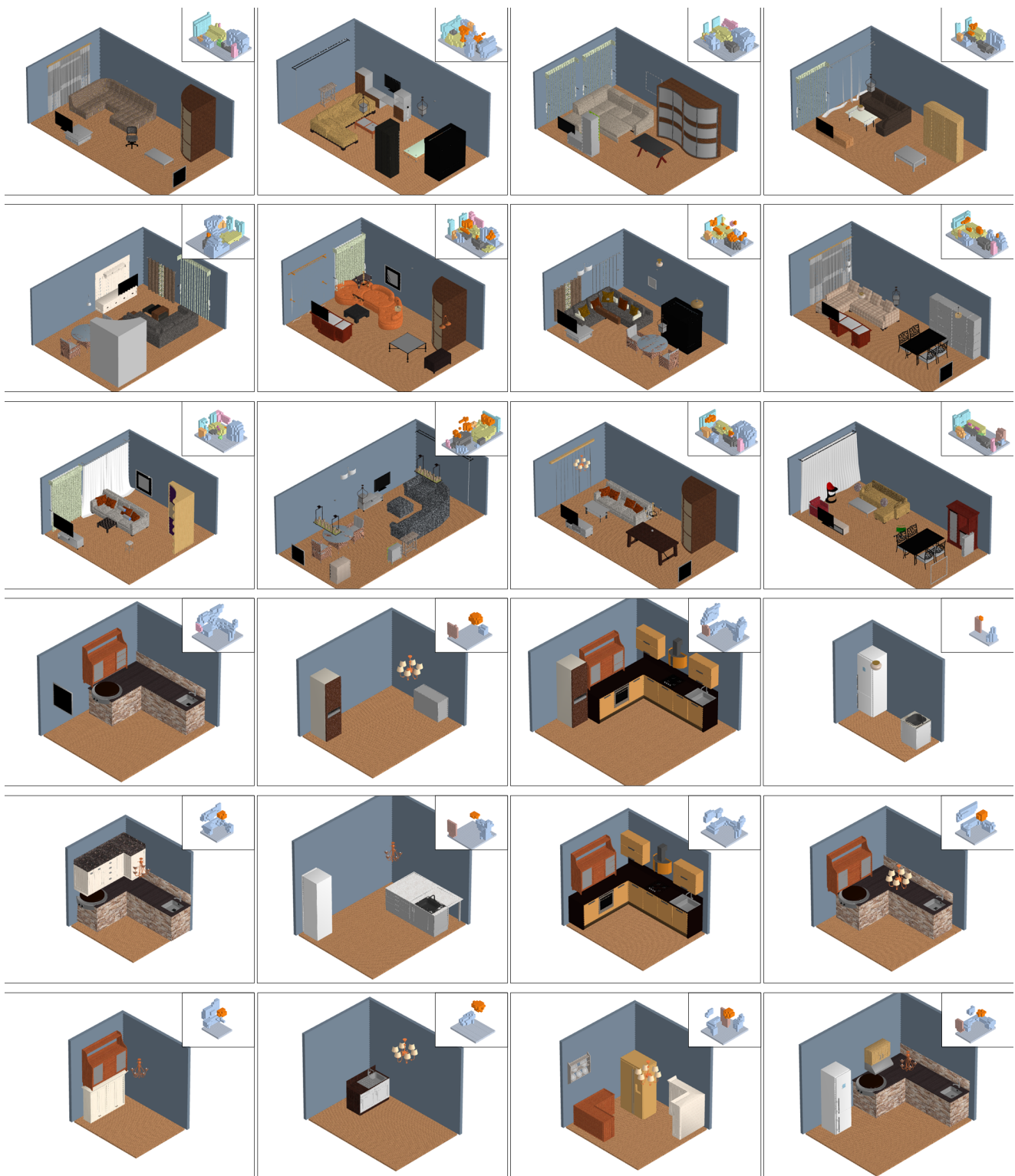

Figure 7. More results generated by our method from Structured3D-LivingRoom (the first three rows) and Structured3D-Kitchen (the last three rows).

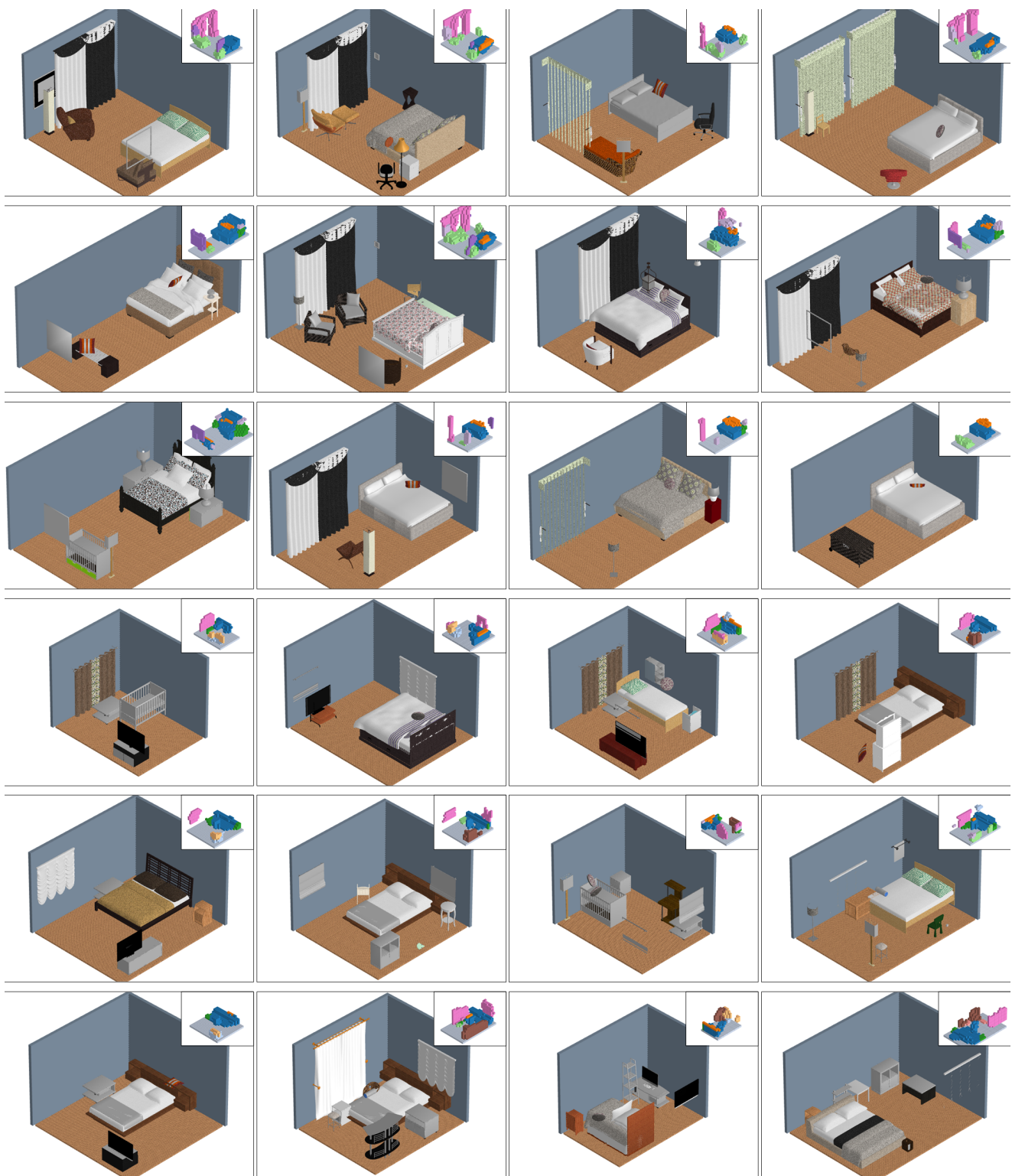

Figure 8. More results generated by our method from Matterport3D-Bedroom (the first three rows) and NYUv2-RGB-Bedroom (the last three rows).



\end{document}